\documentclass[10pt,twocolumn,letterpaper]{article}

\usepackage{iccv}

\usepackage{times}
\usepackage{epsfig}
\usepackage{graphicx}
\usepackage{amsmath}
\usepackage{amssymb}
\usepackage{color}
\usepackage{dsfont}
\usepackage{cite}
\usepackage{float}
\usepackage{caption}
\usepackage{wrapfig}
\usepackage{subfigure}
\usepackage{lipsum}

\usepackage[pagebackref=true,breaklinks=true,letterpaper=true,colorlinks,bookmarks=false]{hyperref}

\iccvfinalcopy 

\def\iccvPaperID{4153} 

\ificcvfinal\fi

\title{\vspace{-5mm}Multiview Supervision By Registration}

\author{Yilun Zhang\\
University of Pennsylvania\\
{\tt\small zhyilun@seas.upenn.edu}
\and
Hyun Soo Park\\
University of Minnesota\\
{\tt\small hspark@umn.edu}
}

\begin{document}

\twocolumn[{%
\maketitle
\begin{center}
\vspace{-8mm}
		\centering
	\includegraphics[width=\textwidth]{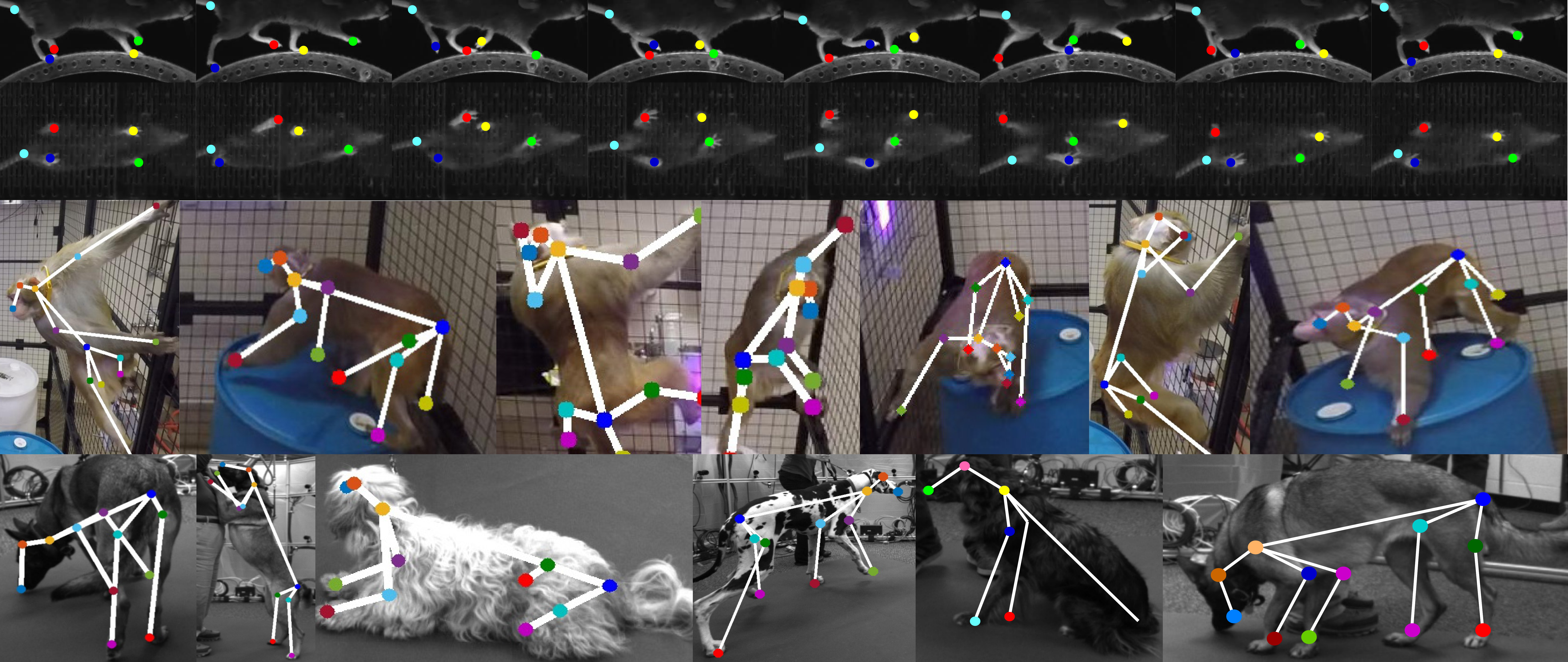}
	\vspace{-5mm}
	\captionof{figure}{This paper presents a semi-supervised keypoint detector by leveraging epipolar geometry and tracking as supervisionary signals. Our approach precisely detects the keypoints of realworld non-human species such as mouse, monkey, and dogs, for which limited labeled data are available ($<4\%$) without a pre-trained model.}
	\label{fig:teaser}
\end{center}	
\label{fig:teaser_main}
}]

\begin{abstract}
    This paper presents a semi-supervised learning framework to train a keypoint detector using multiview image streams given the limited labeled data (typically $<$4\%). We leverage the complementary relationship between multiview geometry and visual tracking to provide three types of supervisionary signals to utilize the unlabeled data: (1) keypoint detection in one view can be supervised by other views via the epipolar geometry; (2) a keypoint moves smoothly over time where its optical flow can be used to temporally supervise consecutive image frames to each other; (3) visible keypoint in one view is likely to be visible in the adjacent view. We integrate these three signals in a differentiable fashion to design a new end-to-end neural network composed of three pathways. This design allows us to extensively use the unlabeled data to train the keypoint detector. We show that our approach outperforms existing detectors including DeepLabCut tailored to the keypoint detection of non-human species such as monkeys, dogs, and mice.  

\end{abstract}

\section{Introduction}\label{sec:Intro}

Enabling computational measurements of the motor behaviors of animals gives rise to scaling up neuroscientific experiments with an unprecedented precision, leading to deeper understanding of our behaviors (humans). For instance, human surrogate models, such as monkeys and mice, have been studied to identify the neural-behavioral pathway through their {\em free-ranging} activities (including several social interactions), which is largely homologous to humans. While non-invasive markerless motion capture is a viable solution to measure such behaviors, it still remains blind to animal behaviors because of lack of a large-scale annotated dataset unlike human subjects (e.g., MS COCO~\cite{lin:2014} and MPII~\cite{Andriluka:2014}). 
 
Recently, subject-agnostic pose tracking approaches based on deep neural networks such as DeepLabCut~\cite{Mathisetal2018} have shown remarkable generalization power, allowing a \textit{smart} pose interpolation: a pre-trained network based on a generic large image dataset (e.g., ImageNet~\cite{ILSVRC15}) is refined to learn a pose variation from a few hundreds of annotated images in a video, and then, the refined network tracks the poses in the rest video by detection. It is relatively labor-effective (comparing to labeling  millions of images) and resilient to a target, i.e., the keypoints on body, foot, and finger of cheetah, insects, and mouse can be reliably tracked. However, their application to the free-ranging behaviors\footnote{Their approaches are designed to track restricted motion, e.g., the animal's head be immobile and attached to a recording rig~\cite{Velliste:2008}.} is challenging because such motion introduces a larger pose variation and self-occlusion, and therefore, considerable amount of annotations is needed. Figure~\ref{fig:PCK_ablation}(d-e) illustrates its performance degradation as the range of motion increases (i.e., mice $\ll$ monkeys). 


This paper presents a new semi-supervised learning approach for a pose detector that leverages the complementary relationship between multiview geometry and visual tracking given the limited labeled data. We hypothesize that the annotation efforts can be substantially reduced by utilizing three self-supervisionary signals embedded in multiview image streams\footnote{Similar insight has been used to reconstruct a reliable long-term 3D trajectories with the multiview videos~\cite{furukawa:2008,joo_cvpr_2014,yoon20173d}.}.
(1) Multiview supervision: the pose detection from two views must satisfy the epipolar constraint, i.e., the detected keypoint in one view must lie in the corresponding epipolar line transferred from the other view given their fundamental matrix~\cite{hartley:2004}. We integrate the cross-view supervision~\cite{jafarian2018monet} by matching the keypoint distributions from two views via their common epipolar plane. This eliminates the necessity of 3D reconstruction\footnote{This is analogous to the fundamental matrix computation without 3D estimation~\cite{longuet-higgins:1981,hartley:2004}.}. (2) Temporal supervision: a pose changes continuously. We incorporate the dense tracking to warp the keypoint distribution between consecutive frames to supervise them to each other~\cite{dong2018supervision,xiaolong:2015}. (3) Visibility supervision: free-ranging activities inherently involve with frequent self-occlusions, producing spurious and degenerate detection. Inspired by the observation that the keypoint visibility varies smoothly across views~\cite{joo_cvpr_2014}, we use the spatial proximity of the cameras to supervise the visibility map in one view from the adjacent views. These three supervisionary signals are combined to form an end-to-end system that effectively uses both labeled and unlabeled data. 

Our system takes as input multiview image streams with a small set of annotated frames, and outputs a pose detection network that predicts the keypoint locations on the rest unlabeled data. We propose a new formulation of multiview semi-supervised learning by matching keypoint distributions conditioned on a visibility map across frames and views. The formulation is implemented using a novel network design composed of three pathways that can minimize the distribution mismatches in the form of four losses: label loss, cross-view loss, tracking loss, and visibility loss. We demonstrate that the resulting network shows strong performance in terms of the keypoint detection accuracy in the presence of significant occlusion given a small set of labeled data ($<$4\%).

Our approach inherits the flexible nature of epipolar geometry, which can be applied to various camera configurations. The distribution matching through their fundamental matrix eliminates the requirement of 3D reconstruction that involves with alternating reconstruction~\cite{simon:2017,Vijayanarasimhan:2017,Byravan:2016,bertasius:2016_unsupervised} or data driven depth prediction~\cite{kanazawaHMR18,drcTulsiani17,zhou2017unsupervised}. Finally, our design is network-agnostic, i.e., any pose detection network producing a probability map representation can be used with a trivial modification such as 
DeepPose~\cite{toshev2014deeppose}, CPM~\cite{cao2016realtime,wei2016convolutional}, and Hourglass~\cite{newell2016stacked}.





To our knowledge, this is the first paper that leverages the spatiotemporal relationship of multiview image streams to train a pose detector. The core contributions include: (1) a new differentiable formulation of multiview spatiotemporal self-supervision for the unlabeled data; (2) a visibility supervision based on camera spatial proximity to prevent from spurious propagation of the self-supervision; (3) its realization using an end-to-end network that is flexible to camera configurations; and (4) strong performance on the realworld data of non-human species on monkeys, dogs, and mice with a small set of the labeled data.

\section{Related Work} \label{sec:related}
This paper studies designing a pose detector given the limited labeled data by leveraging multiview epipolar geometry and temporal consistency. These two supervisionary signals are by large studied in isolation. 


\noindent\textbf{Temporal Supervision} The tracking results such as optical flow~\cite{baker2004lucas}, MOSSE \cite{bolme2010visual}, and discriminative correlation filters~\cite{henriques2015high}, provides an auxiliary information that can be used to enforce the temporal consistency across a continuous sequence~\cite{dong2018supervision,xiaolong:2015}. A challenge is that it suffers from tracking drift induced by object deformation, which substantially limits its validity. Such challenge has been addressed by learning the temporal evolution of tracking patches~\cite{liu2018two,peng2016recurrent} using recurrent neural networks. This generates a compromised network that minimizes the inconsistency in the learned trajectories, which suppresses the low-quality detection from the tracking drift. A pitfall of this approach is the requirement of per-frame annotation to supervise the recurrent network. This requirement can be relaxed by using supervision-by-registration approach~\cite{dong2018supervision} that achieves higher detection rate even with the limited labeled data. However, its application towards the pose detection for non-human species is still challenging because: (1) supervision from optical flow involves with the tracking drift caused by occlusion, and therefore, long-term tracking is infeasible; (2) the soft-argmax operation for computing
the track coordinate may lead to noisy supervision in the cases where the pose detection is erroneous (e.g., multiple peaks) as shown in Figure~\ref{fig:soft_argmax}. This multi-modality of pose recognition escalates when the keypoint is invisible. This strongly influences tracking accuracy, especially for a small-sized target; (3) the argmax operation takes into account only for the peak location where the non-maximum local peaks may play a role.

\noindent\textbf{Multiview Supervision} Multiview images possess highly redundant yet distinctive visual information that can be used to self-supervise the unlabeled data. Bootstrapping is a common practice: to use multiview images to robustly reconstruct the geometry using the correspondences and to project to the unlabeled images to provide a pseudo-label, which has been shown highly effective~\cite{simon:2017,Vijayanarasimhan:2017,Byravan:2016}. A pitfall of this approach is that it involves with an iterative process over learning and reconstruction. Another approach is to separately learn depth from a single view image in isolation that can be used for self-supervision~\cite{kanazawaHMR18,drcTulsiani17,zhou2017unsupervised}. This relies on the depth prediction where the accuracy of the trained model is bounded by the accuracy of reconstruction/prediction. Jafarian et al.~\cite{jafarian2018monet} introduces a new framework that bypasses 3D reconstruction during the training process through the epipolar constraint, i.e., the epipolar constraint is transformed to the distribution matching. The problem of this approach is that its performance is highly dependent on the pre-trained model. It has no reasoning about outliers, i.e., the recognition network converges to a trivial solution if the outliers dominate the distribution of the multiview pose detection.  

Our main hypothesis is that these two supervisions are complementary. We formulate the spatiotemporal supervision that can benefit from both and address each limitation. (1) We use dense optical flow tracking to address noisy supervision, i.e., it is unlikely that the noisy prediction is temporally correlated. (2) We leverage the end-to-end epipolar distribution matching to avoid the multimodality issue that arises using the soft-argmax operation. This is differentiable, and therefore, trainable. (3) The multiview image streams can alleviate the tracking drift~\cite{joo_cvpr_2014,yoon20173d}, i.e., it is unlikely that the tracking drift occurs in a geometrically consistent fashion. (4) Visibility map can assist to determine the validity of the tracking without explicit outlier rejection. 

\section{Notation and Multiview Conditions} \label{Sec:notation}

\begin{figure}[t]
\begin{center}
\subfigure[Spurious soft-argmax]{\label{fig:soft_argmax}\includegraphics[height=0.17\textwidth]{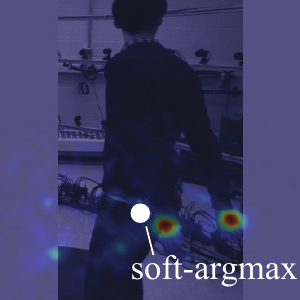}}~~~~~
\subfigure[Multiview supervision]{\label{Fig:geom}\includegraphics[height=0.2\textwidth]{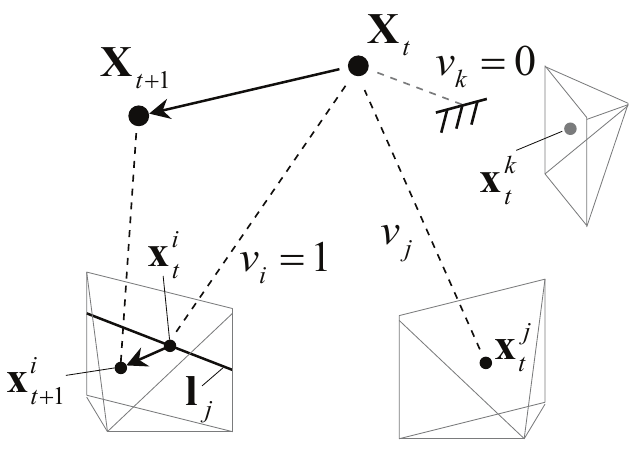}}
\end{center}
\vspace{-5mm}
   \caption{(a) Soft-argmax produces a biased keypoint estimate when the keypoint distribution is multimodal. (b) We use three self-supervisionary signals: cross-view supervision ($\mathbf{l}_j^\mathsf{T}\widetilde{\mathbf{x}}_t^j$), temporal supervision ($\mathbf{x}_t^i=W_{t+1\rightarrow t}(\mathbf{x}_{t+1}^i)$), and visibility supervision ($v_i\approx v_j$).  }
\label{Fig:ge}
\vspace{-5mm}
\end{figure}

Consider multiview image streams, $\mathcal{I} = \{\mathbf{I}_t^i\}$ where $\mathbf{I}_t^i$ is the image of the $i^{\rm th}$ camera at $t$ time instant. We denote the set of synchronized images at $t$ time instant across all views with $\mathcal{I}_t = \{\mathbf{I}_t^1,\cdots\mathbf{I}_t^n\}$ that satisfy the epipolar constraint~\cite{longuet-higgins:1981} where $n$ is the number of cameras. $\mathcal{I}^i = \{\mathbf{I}_1^i,\cdots\mathbf{I}_T^i\}$ is the set of images from the $i^{\rm th}$ camera for all time instances where $T$ is the total time instances\footnote{We consider a stationary multi-camera system~\cite{joo_cvpr_2014,yoon20173d} while the spatiotemporal constraint of epipolar geometry and temporal coherence still applies for a moving synchronized multi-camera system, e.g., social cameras~\cite{arev:2014}.}.  A subset of these images are manually annotated (keypoint location) $\mathcal{I}_L$, and the rest remain unlabeled $\mathcal{I}_U$, i.e., $\mathcal{I} = \mathcal{I}_L \cup \mathcal{I}_U$.




\begin{figure*}[t]
\begin{center}
\subfigure[Epipolar plane]{\label{Fig:crss_geom}\includegraphics[height=0.19\textwidth]{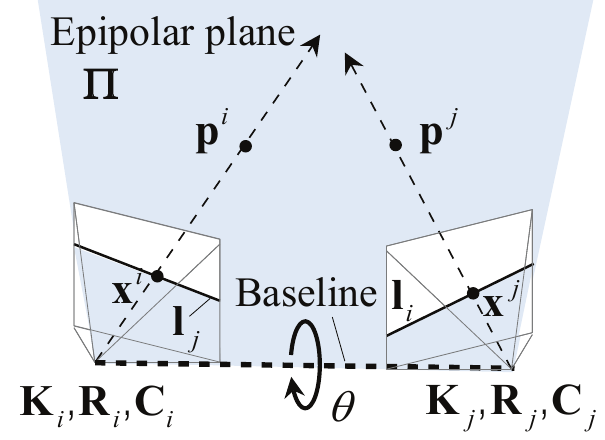}}~~~~~~~
\subfigure[Cross-view supervision]{\label{Fig:mouse_transfer}\includegraphics[height=0.19\textwidth]{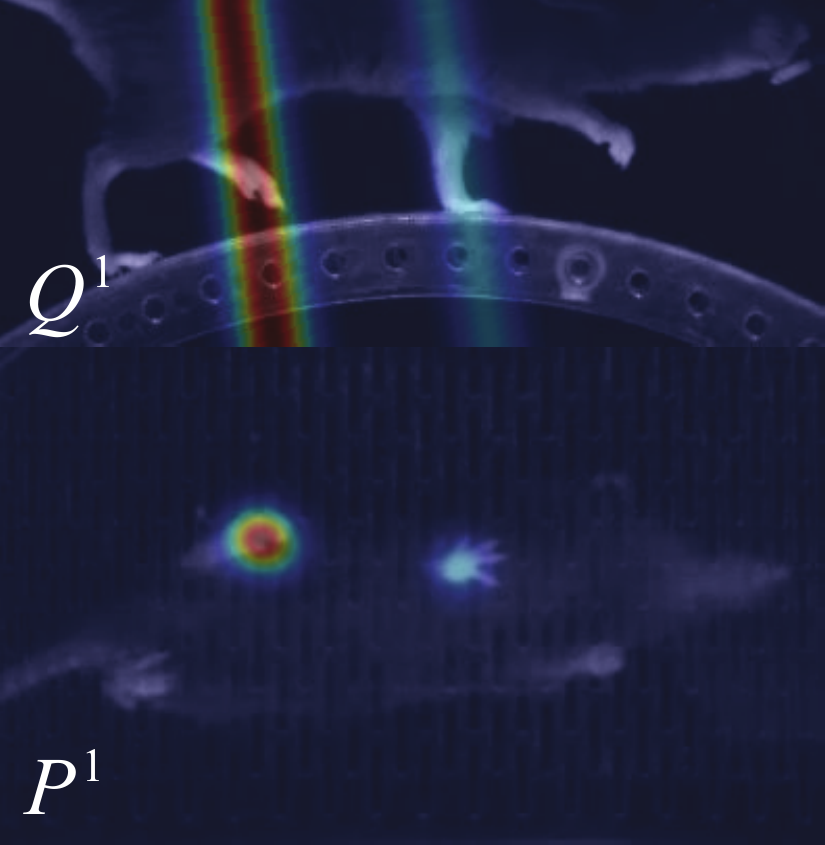}}~~~~~~~
\subfigure[Temporal supervision]{\label{Fig:tempor_transfer}\includegraphics[height=0.19\textwidth]{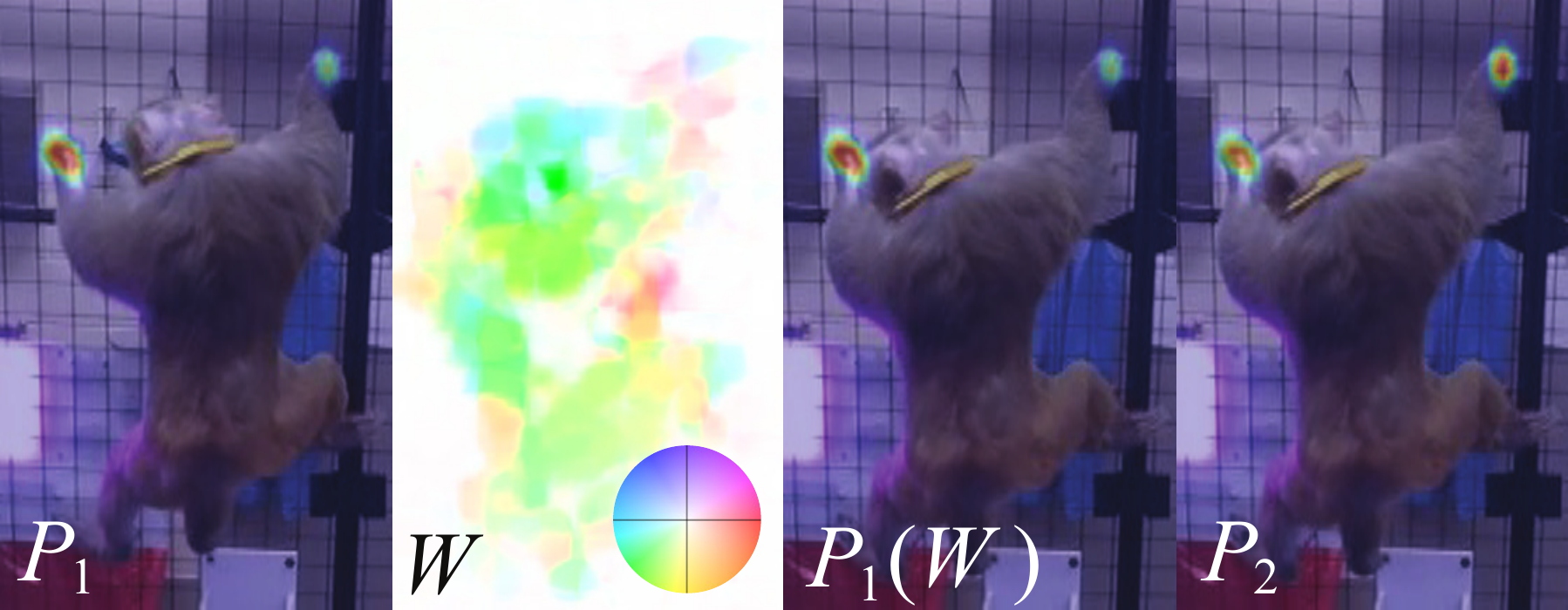}}
\end{center}
\vspace{-5mm}
   \caption{(a) A keypoint distribution can be transformed to the common epipolar plane distribution, allowing cross-view supervision. (b) The keypoint distribution of the hind right foot in view 1 ($P^1$) is transformed to $Q^i$ and projected onto the side view (top). (c) The keypoint distribution can be warped $P_1(W)$ using dense optical flow ($W$) to supervise the next frame $P_2$, which is multimodal distribution.}
\label{Fig:ge}
\vspace{-5mm}
\end{figure*}


A 3D keypoint $\mathbf{X}_t\in\mathds{R}^3$ at $t$ time instant travels to $\mathbf{X}_{t+1}$. The point is projected onto the $i^{\rm th}$ and $j^{\rm th}$ images ($\mathbf{I}^i_t$ and $\mathbf{I}^j_t$) to form the 2D projections $\mathbf{x}_t^i, \mathbf{x}_t^j\in\mathds{R}^2$ as shown in Figure~\ref{Fig:geom}:
\begin{align}
    \widetilde{\mathbf{x}}_t^i \cong \mathbf{P}^i \widetilde{\mathbf{X}}_t,~~~~~\widetilde{\mathbf{x}}_t^j \cong \mathbf{P}^j \widetilde{\mathbf{X}}_t,~~~~~\widetilde{\mathbf{x}}_{t+1}^i \cong \mathbf{P}^i \widetilde{\mathbf{X}}_{t+1}, \label{Eq:multi}
\end{align}
where $\mathbf{P}^i \in \mathds{R}^{3\times 4}$ is the $i^{\rm th}$ camera projection matrix, and $\widetilde{\mathbf{x}}$ is the homogeneous representation of $\mathbf{x}$~\cite{hartley:2004}. 

To be geometrically consistent across multiview image streams, the projections of the moving 3D keypoint need to satisfy the following three constraints: \\
\noindent \textbf{Cross-view Constraint} The keypoint $\mathbf{x}_t^i$ must lie in the epipolar line of the corresponding point $\mathbf{x}_t^j$ in the $j^{\rm th}$ view~\cite{hartley:2004}, i.e., $(\widetilde{\textbf{x}}_t^j)^\mathsf{T} \mathbf{F}_{ij} \widetilde{\mathbf{x}}_t^i = \mathbf{l}_j^\mathsf{T}\widetilde{\textbf{x}}_t^i = 0$ where $\mathbf{F}_{ij}$ is the fundamental matrix between the $i^{\rm th}$ and $j^{\rm th}$ views, and $\mathbf{l}_j \in \mathds{P}^2$ is the epipolar line transferred from $\mathbf{x}_t^j$. \\
\noindent \textbf{Tracking Constraint} The pixel brightness on $\mathbf{x}_t$ and $\mathbf{x}_{t+1}$ must be persistent, $\mathbf{I}_t^i(\mathbf{x}_{t+1}^i+\Delta \mathbf{x}) = \mathbf{I}_{t+1}^i(\mathbf{x}_{t+1}^i)$ where $\Delta \mathbf{x}$ is the backward optical flow at $\mathbf{x}_{t+1}^i$. \\
\noindent \textbf{Visibility Constraint} The visible keypoint in one view is likely visible in adjacent view, i.e., $v_i\approx v_j$ if $\|\mathbf{C}_i-\mathbf{C}_j\|<\epsilon$ where $v_i \in [0,1]$ is the probability of the keypoint being visible to the $i^{\rm th}$ camera, and $\mathbf{C}_i$ is the optical center of the $i^{\rm th}$ camera. For instance, $v_i=v_j=1$ and $v_k=0$ in Figure~\ref{Fig:geom}. 

\section{Multiview Supervision by Registration} \label{Sec:MSBR}

We build a keypoint detector producing the keypoint distribution $\phi(\mathbf{I};\mathbf{w})\in [0,1]^{W\times H \times C}$ and its visibility map $\psi(\mathbf{I};\mathbf{w}_v)\in [0,1]^{W\times H \times C}$. These two distributions are combined to produce a posterior per-pixel keypoint distribution:
\begin{align}
 \xi(\mathbf{I}) = \phi(\mathbf{I};\mathbf{w}) \psi(\mathbf{I};\mathbf{w}_v)   \label{Eq:posterior}
\end{align}
where $W$, $H$, and $C$ are the width, height, and the number of keypoints including the background. The keypoint distribution is parametrized by the weight $\mathbf{w}$, and the visibility map is parametrized by the weight $\mathbf{w}_v$. We denote the probability evaluated at $\mathbf{x}$ as $P_t^i(\mathbf{x}) = \left. \phi(\mathbf{I}_t^i;\mathbf{w}) \right|_{\mathbf{x}}$ and $V_t^i(\mathbf{x}) = \left. \psi(\mathbf{I}_t^i;\mathbf{w}_v) \right|_{\mathbf{x}}$. In the inference phase, the resulting detected keypoint location is the peak in the posterior distribution $\xi$.

We learn $\mathbf{w}$ and $\mathbf{w}_v$ from the labeled and unlabeled data where $|\mathcal{I}_L| \ll |\mathcal{I}_U|$ where a supervised learning approach alone likely to be highly biased. To utilize the unlabeled data, we leverage the three multiview constraints in Section~\ref{Sec:notation}. However, integrating these into an end-to-end training is challenging because of \textit{representation mismatch}. The raster representation of the keypoint distribution $\phi(\mathbf{I};\mathbf{w})$ differs from the vector representation of the constraints (e.g., $\mathbf{l}^\mathsf{T}\widetilde{\mathbf{x}}=0$). Conversion between these two representations requires the argmax operation:
\begin{align}
    \mathbf{x}^* = \underset{\mathbf{x}}{\operatorname{argmax}} ~~P_t^i(\mathbf{x}). \label{Eq:argmax}
\end{align}
The argmax in Equation~(\ref{Eq:argmax}) is non-differentiable, and therefore, embedding the constraints makes the network not trainable. This precludes from an end-to-end training for multiview supervision, leading to offline alternating reconstruction~\cite{simon:2017,Vijayanarasimhan:2017,Byravan:2016,bertasius:2016_unsupervised} or additional depth prediction~\cite{kanazawaHMR18,drcTulsiani17,zhou2017unsupervised} that often suffer from suboptimality~\cite{jafarian2018monet}. Whilst the differentiable soft-argmax can alleviate this issue to some extent, it is highly sensitive to spurious and noisy keypoint detection (e.g., multimodal probability map as shown in Figure~\ref{fig:soft_argmax}). In subsequent sections, we address this challenge by transforming the constraints into a distribution matching with the raster representation as a whole by minimizing KL divergence~\cite{Kullback:1951}.

\subsection{Cross-view Supervision}
A set of images at the same time instant, $\mathcal{I}_t$, we supervise their keypoint distributions based on the epipolar constraint. Inspired by Jafarian et al.~\cite{jafarian2018monet}, we reformulate the epipolar geometry in terms of distribution matching over their common epipolar planes. Consider a keypoint in the $i^{\rm th}$ image, $\mathbf{x}^i$, that corresponds to the keypoint in the $j^{\rm th}$ image $\mathbf{x}^j$. Their inverse projections (the 3D ray emitted from the camera center and passing the keypoint location $\mathbf{x}_i$) can be written as $\mathbf{p}_i(\lambda)=\lambda \mathbf{R}_i^\mathsf{T}\mathbf{K}_i^{-1}\widetilde{\mathbf{x}}_i + \mathbf{C}_i$ where $\mathbf{K}_i\in\mathds{R}^{3\times3}$, $\mathbf{R}_i\in SO(3)$, and $\mathbf{C}_i\in\mathds{R}^3$ are the intrinsic parameter, rotation, and optical center of the $i^{\rm th}$ camera, and $\lambda>0$ is the depth of the point on the ray as shown in Figure~\ref{Fig:crss_geom}. To satisfy the epipolar constraint, their inverse projections must lie in a common epipolar plane ($\boldsymbol{\Pi}\in \mathds{P}^3$), i.e., $\boldsymbol{\Pi}^\mathsf{T}\widetilde{\mathbf{p}}_i=\boldsymbol{\Pi}^\mathsf{T}\widetilde{\mathbf{p}}_j = 0$. 

Using the fact that the common epipolar plane can be parametrized by its rotation about the baseline, i.e., surface normal $\boldsymbol{\Pi}(\theta \in \mathds{S})$, we transform the keypoint distribution to the epipolar plane distribution, obtained by the max-pooling over the epipolar line:
\begin{align}
    Q^i(\theta) = \underset{\mathbf{x} \in \mathbf{l}_j(\theta)}{\operatorname{argmax}} ~~P^i(\mathbf{x}), \label{Eq:cross-sup}
\end{align}
where $\mathbf{l}_j(\theta)$ is the epipolar line that is the projection of the common epipolar plane, and $Q^i$ is the epipolar plane distribution. See Appendix for more details. The bottom row in Figure~\ref{Fig:mouse_transfer} illustrates the keypoint distribution of the right hind foot in view 1 ($P^1$). It is transformed to the epipolar plane distribution $Q^1$ using the max-pooling over the epipolar lines. We visualize the projection of $Q^1$ onto the second view (the top row), i.e., the hind foot must lie in the most probable location in the second view. Note that the multimodal keypoint distribution does not produce additional spurious supervision to the other view.

Equation~(\ref{Eq:cross-sup}) allows measuring geometric discrepancy of keypoint distributions across views. Therefore, the unlabeled data can be self-supervised to each other by minimizing their cross entropy with the raster representation:
\begin{align}
    L_{\rm C}(\mathcal{I}_t) = \sum_{i,j \in \mathcal{C}} D_\mathrm{KL} (Q_i||Q_{j}), \label{Eq:mcs}
\end{align}
where $\mathcal{C}$ is the camera index set of $\mathcal{I}_t$.

\subsection{Temporal Supervision}
Given a sequence of images from the $i^{\rm th}$ camera, $\mathcal{I}^i$, we supervise the keypoint distribution at $t^{\rm th}$ time instant using that of neighboring images in time, i.e.,
\begin{align}
    P^i_{t_1}(\mathbf{x}) \approx P^i_{t_2}(W_{t_2\rightarrow t_1}(\mathbf{x})) \label{Eq:temporal}
\end{align}
where $W_{t_2\rightarrow t_1}$ is the pre-computed dense optical flow from $t_2$ to $t_1$ frames, i.e., $P^i_{t_2}(W_{t_2\rightarrow t_1}(\mathbf{x}))$ is the warped distribution of $P^i_{t_2}$. We use a kernelized correlation filter~\cite{henriques2015high} with inverse compositional mapping~\cite{baker2004lucas} to track all pixels offline while online optical flow computation~\cite{dong2018supervision, lin:2017} can be complementary to our approach with a trivial modification. 

Using Equation~(\ref{Eq:temporal}), we design a tracking loss for the temporal supervision:
\begin{align}
    L_{\rm T}(\mathcal{I}^i) = \sum_{t_1,t_2 \in [0,T]}  D_{\rm KL}(P^i_{t_1}||P^i_{t_2}(W_{t_2\rightarrow t_1})), \label{Eq:temporal_sup}
\end{align}
where $T$ is the number of frames. 

A key innovation of Equation~(\ref{Eq:temporal_sup}) against existing optical flow supervision~\cite{dong2018supervision,xiaolong:2015,lin:2014} is that it eliminates the necessity of the argmax operation by warping the keypoint distribution as a whole. In practice, we find that having sufficient time difference between frames improves training performance and efficiency. For instance, a high framerate video of a monkey who stays still for a majority of time generates less informative temporal supervision and is prone to noise, i.e., $W_{t+1\rightarrow t} = I$ where $I$ is the identity mapping. On the other hand, when the frame difference is too large, significant tracking drift is likely to occur. We address this by selectively applying the temporal supervision on the two frames that have the sufficient magnitude of the integral dense optical flow, i.e., $\epsilon_m < \sum_{x\in \mathcal{X}} \|W_{t_2\rightarrow t_1}(\mathbf{x})\|<\epsilon_M$ where $\mathcal{X}$ is the domain of an image, and $\epsilon_m$ and $\epsilon_M$ are lower and upper bounds of the magnitude of the integral dense optical flow. Figure~\ref{Fig:tempor_transfer} illustrates the temporal supervision using dense optical flow. The left wrist keypoint distribution $P_{1}$ is warped to form $P_{1}(W)$. This unimodal distribution can supervise the ambiguous prediction in $P_2$ with two modes. 

\subsection{Visibility Supervision}

\begin{figure}
    \centering
    \includegraphics[width=0.47\textwidth]{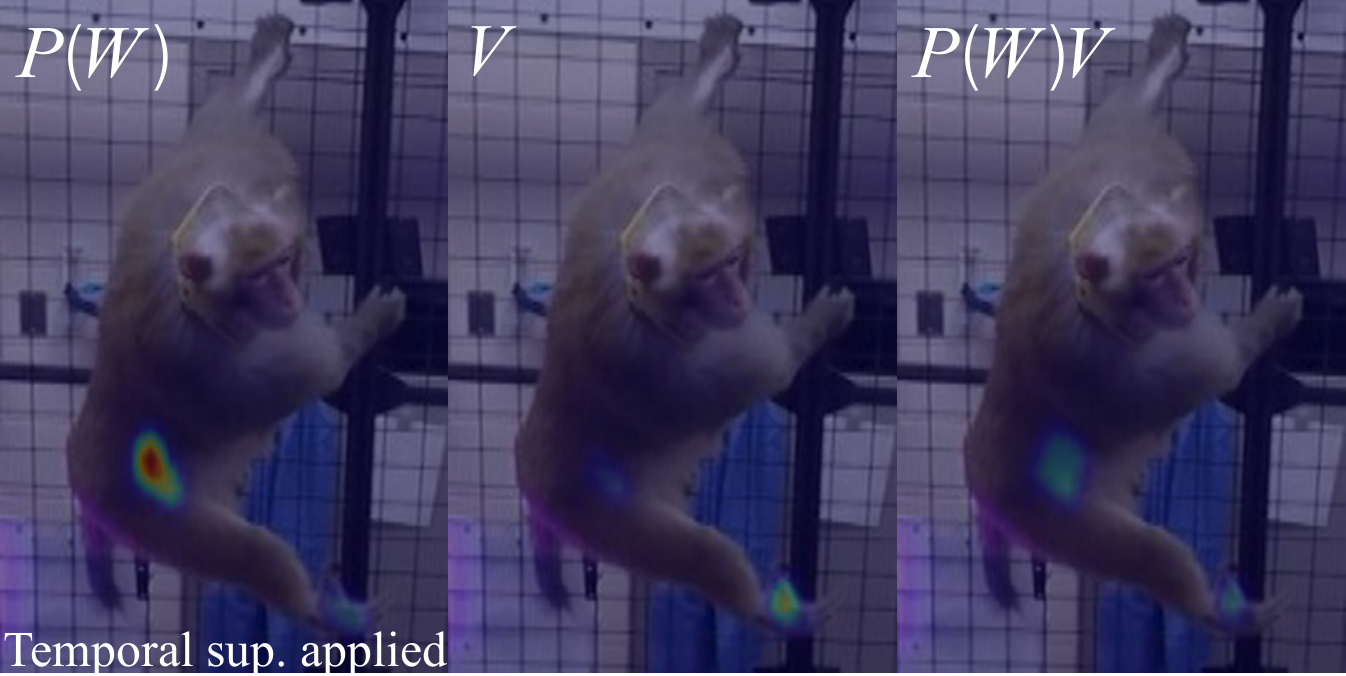}
    \caption{We integrate a visibility inference to validate the multiview supervisionary signals. The left hind paw is occluded by torso, which is conditioned by the visibility map (middle), resulting in the reduction of the keypoint probability. This prevents from influencing the occluded keypoint detection across views.}
    \label{fig:visibility}
\end{figure}

Free-ranging activities inherently involve with self-occlusion, e.g., a hand is occluded by the torso at a certain view. Without precise reasoning about the visibility of keypoints, the cross-view and temporal supervisions can be highly fragile because there is no mechanism to prevent from such error propagation over the unlabeled multiview images\footnote{A similar observation has been made for long-term trajectory reconstruction~\cite{joo_cvpr_2014}.}. For instance, the temporal supervision via the optical flow of the occluded hand can mislead the hand location to the torso location in other visible images. To reject such error, RANSAC~\cite{Fischler:1981} with geometric verification (e.g., reprojection error) has been used. However, the operation is non-differentiable, and therefore, it requires alternating offline reconstruction and training~\cite{simon:2017}. 

\begin{figure*}[t]
  \centering  
      \includegraphics[width=\textwidth]{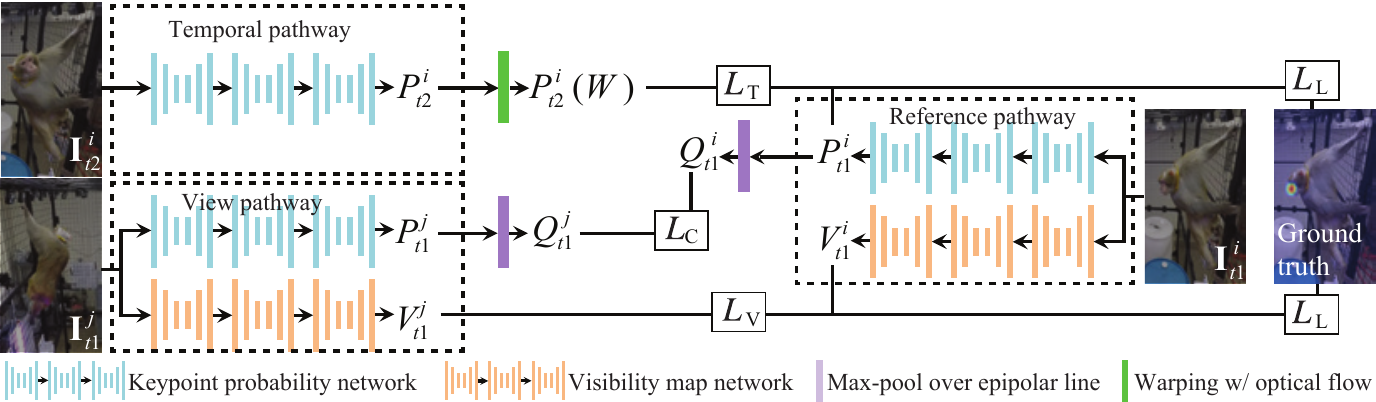}
  \caption{We design a network composed of three pathways: reference, temporal, and view pathways to utilize both labeled and unlabeled data. Each pathway is composed of two subnetworks producing keypoint distribution and visibility map. The labeled loss $L_{\rm L}$ is computed from the reference pathway by comparing to the ground truth annotation (keypoint and visibility) if available. The temporal and reference pathways measure the tracking loss $L_{\rm T}$ by warping the keypoint distribution using the dense optical flow ($P_{t_2}^i(W_{t_2\rightarrow t_1})$), and the view and reference pathways measure the cross-view loss $L_{\rm C}$ by transforming the keypoint distribution to the epipolar plane distribution, i.e., $Q_{t_1}^i\leftrightarrow Q_{t_1}^j$. } \label{Fig:network}
\end{figure*}


Instead, we design a new module that integrates the visibility inference as a part of the training process. The key idea is that a keypoint is likely to be visible if it is visible from the adjacent cameras. This provides a spatial prior on the visibility map across views:
\begin{align}
    L_{\rm V}(\mathcal{I}_t) = \sum_{i,j \in \mathcal{C}} \delta_{i,j}\|\max~V_t^i - \max~V_t^j\|^2, \label{Eq:visi}
\end{align}
where $\delta_{i,j}$ is Kronecker delta that is one if the distance between the optical centers of the $i^{\rm th}$ and $j^{\rm th}$ cameras is smaller than $\epsilon_C$, i.e., $\|\mathbf{C}_i-\mathbf{C}_j\| < \epsilon_C$, and zero otherwise, and $\mathcal{C}$ is the camera index set of $\mathcal{I}_t$. Equation~(\ref{Eq:visi}) is a necessary condition that penalizes the difference in visibility maps for adjacent cameras, i.e., it is valid when the location of the maximum visibility map coincides with the peak of the keypoint probability. In practice, the visibility is highly correlated with the keypoint distribution where $L_{\rm V}$ is effective. For instance, Figure~\ref{fig:visibility} illustrates the visibility supervision across views. The left hind paw is occluded by torso, which is conditioned by the visibility map (middle), resulting in the reduction of the keypoint probability. This prevents from influencing the occluded keypoint detection across views.

\subsection{Label Supervision}
We supervise the keypoint distribution and visibility map using a set of the labeled data as follows:
\begin{align}
    L_{\rm L}(\mathcal{I}_L) = \sum_{\mathbf{I}\in \mathcal{I}_L} D_{\rm KL} (\overline{P}_{\mathbf{I}_{t}^i}||P_t^i) + D_{\rm KL} (\overline{V}_{\mathbf{I}_{t}^i}||V_t^i), \label{Eq:labelel}
\end{align}
where $\overline{P}_{\mathbf{I}_{t}^i}$ and $\overline{V}_{\mathbf{I}_{t}^i}$ are the ground truth keypoint distribution and its visibility of image $\mathbf{I}_t^i$. The ground truth keypoint distribution is obtained by convolving a scaled Gaussian at the ground truth keypoint location. For the visibility map, it is computed via ray-casting on a discretized 3D voxel space. See Appendix for more details of ground truth visibility map generation.


\subsection{Overall Loss}
The resulting keypoint detector is learned using both labeled and unlabeled data by minimizing the following overall loss:
\begin{align}
   L(\mathbf{w}, \mathbf{w}_v) &= L_{\rm L} (\mathcal{I}) + \lambda_{\rm C} \sum_{t=1}^T L_{\rm C} (\mathcal{I}_t) + \lambda_{\rm T} \sum_{i \in \mathcal{C}} L_{\rm T}(\mathcal{I}^i)\nonumber\\ &+ \lambda_{\rm V} \sum_{t=1}^T L_{\rm V} (\mathcal{I}_t), 
\end{align}
where $L_{\rm L}$, $L_{\rm C}$, and $L_{\rm T}$, and $L_{\rm V}$ are the losses for the labeled supervision, cross-view supervision, temporal supervision, and visibility supervision, respectively. $\lambda_C$, $\lambda_T$ and $\lambda_V$ are the weights that control their importance.

\section{Implementation}
We design a network that is composed of three pathways: reference, view, and temporal pathways as shown in Figure~\ref{Fig:network}. Each pathway takes as an input image with the size of $368\times368\times3$ and produce the keypoint probability and visibility map with the size of $46\times46\times 21$. They all share the network weights $\mathbf{w}$ and $\mathbf{w}_v$. The reference and view pathways are designed to measure the cross-view loss $L_{\rm C}$ and visibility supervision loss $L_{\rm V}$ for two adjacent views by transforming to the epipolar plane distribution. The reference and temporal pathways measure the tracking loss $L_{\rm T}$ by warping the keypoint distribution using the dense optical flow. The label loss is measured for the reference pathway if the input image is labeled. We use the convolutional pose machine~\cite{cao2016realtime} as a base network to implement $\phi(\cdot)$ and $\psi(\cdot)$ while any existing pose detector can be complementary. See Appendix for network training. The code is publicly available: \href{https://github.com/msbrpp/MSBR}{https://github.com/msbrpp/MSBR}


\noindent\textbf{Network Initialization by Bootstrapping} To alleviate the noisy initialization of the detector, which occurs frequently when the unlabeled data dominate, we take a few practical steps. (1) With a subset of the labeled data in the same time instant, we triangulate the keypoint in 3D with RANSAC. This 3D keypoint is projected onto all multiview images, which can greatly augment the labeled data reliably. (2) Based on the 3D keypoints with volume estimation, we compute the visibility using ray-casting, which provides the visibility map label for all views. (3) With the augmented labeled data with their visibility, we train the network in a fully supervised manner. This process is called bootstrapping~\cite{simon:2017}, which provides a good initialization to train our triple network. (4) We re-train the pre-trained network with the unlabeled data with cross-view, tracking, and visibility losses. 

\begin{table*}[t]
\scriptsize
\begin{center}
\begin{tabular}{l|cccc|c||cccc|c||ccccc|c}
\hline
&\multicolumn{5}{c||}{Human subject I} & \multicolumn{5}{c||}{Dog subject} &\multicolumn{6}{c}{Monkey subject}\\
\hline
Method & Sho & Elb & Wri & Kne  &\textbf{AUC} 
  & Nec & F.Leg &  Paw  &H. Leg&\textbf{AUC}
          & Nec           & F.Leg         & Paw           & Hip           & H. Leg        & \textbf{AUC}\\
\hline
Supervised learning                          
&  81.7 & 37.9 & 33.6  & 86.1  & 91.6
 & 96.1 & 80.3 &34.8  & 82.1  & 91.3
            & 94.5          & 67.4          & 31.5          & 96.9          & 68.9          & 75.3\\
Temp.  
& 86.4 & 44.6 & 32.5 & 93.4 & 91.7
 & 94.2 & 83.2 &31.6 & 83.3  & 92.0
         & 94.2          & 82.8          & 37.4          & 90.3          & 83.7          & 87.4\\
Temp. + Vis.                        
& 92.7 & 48.4 & 41.1  & 97.8  & 93.3
 & 96.9 & 91.5 &38.1  & 88.9   & 92.5
         & 94.9          & 87.4          & 45.8          & 91.6          & 87.9          & 89.2\\
Cross.  
 & 62.4 & 31.7 & 19.8  & 44.7  & 78.7
 & 85.3 & 68.7 &23.6 & 61.4   & 70.3
            & 89.7          & 60.2          & 29.6          & 50.9          & 63.7          & 68.9\\
Cross. + Boot.                       
 & 85.0 & 41.5 & 38.6  & 97.6 & 92.6
 & 96.6 & 88.2 &35.3 & 91.2   & 92.9
          & 94.2          & 87.4          & 38.2          & 91.7          & 86.2          & 87.6 \\
Temp. + Cross.                
 & 88.8 & 70.6 & 40.2  & 97.5  & 92.2 
 & 96.1 & 89.1 &37.2  & 92.3   & 92.9
         & 97.6          & 92.1          & 47.2          & 90.4          & 93.5          & 90.3 \\
Temp. + Cross + Boot.             
 & 89.4 & 77.1 & 57.5 & 98.6  & 92.2
 & 98.9 & 92.5 &52.8  & 95.8   & 93.8
       & 97.9          & 94.8          & 48.7          & 92.0          & 95.1          & 91.6\\
\hline\hline
Ours                                 
& \textbf{92.9 }& \textbf{77.2 }& \textbf{65.4 } & \textbf{98.9 } & \textbf{95.1}
& \textbf{98.9 }& \textbf{94.2}& \textbf{53.2} & \textbf{95.8} &\textbf{94.8}
 & \textbf{98.7} & \textbf{95.2} & \textbf{50.1} & \textbf{93.5} & \textbf{95.7} & \textbf{92.2}\\
\hline

\end{tabular}
\end{center}
\vspace{-5mm}
\caption{We conduct an ablation study on human, dog, and monkey subjects using the PCKh measure.}
\label{tb:pckh}
\end{table*}

\setlength{\tabcolsep}{4pt}
\begin{table*}[t]
\footnotesize
\begin{center}
    
\begin{tabular}{l|cllllll|l||cllllll|l}
\hline
&\multicolumn{8}{c||}{Unlabeled data detection}&\multicolumn{8}{c}{Unseen data detection}\\
\hline
Panoptic Studio dataset                   & Nec                      & \multicolumn{1}{c}{Sho}           & \multicolumn{1}{c}{Elb}           & \multicolumn{1}{c}{Wri}           & \multicolumn{1}{c}{Hip}           & \multicolumn{1}{c}{Kne}           & \multicolumn{1}{c|}{Ank}           & \multicolumn{1}{c||}{\textbf{AUC}} & Nec                      & \multicolumn{1}{c}{Sho}           & \multicolumn{1}{c}{Elb}           & \multicolumn{1}{c}{Wri}           & \multicolumn{1}{c}{Hip}           & \multicolumn{1}{c}{Kne}           & \multicolumn{1}{c|}{Ank}           & \multicolumn{1}{c}{\textbf{AUC}}  \\ \hline
Supervised learning               & 93.5                     & \multicolumn{1}{c}{78.2}          & \multicolumn{1}{c}{36.8}          & \multicolumn{1}{c}{28.6}          & \multicolumn{1}{c}{98.7}          & \multicolumn{1}{c}{83.5}          & \multicolumn{1}{c|}{92.4}          & \multicolumn{1}{c||}{88.5}     
& 94.2                              & \multicolumn{1}{c}{75.4}          & \multicolumn{1}{c}{32.9}          & \multicolumn{1}{c}{23.6}          & \multicolumn{1}{c}{97.2} & \multicolumn{1}{c}{78.6}          & \multicolumn{1}{c|}{89.4}          & \multicolumn{1}{c}{85.5}  \\
Dong et al.~\cite{dong2018supervision}            & 98.1                     & \multicolumn{1}{c}{88.3}          & \multicolumn{1}{c}{43.6}          & \multicolumn{1}{c}{33.5}          & \multicolumn{1}{c}{97.8}          & \multicolumn{1}{c}{92.7}          & \multicolumn{1}{c|}{96.6}          & \multicolumn{1}{c||}{92.3}          
& 96.7                              & \multicolumn{1}{c}{80.7}          & \multicolumn{1}{c}{37.8}          & \multicolumn{1}{c}{28.2}          & \multicolumn{1}{c}{\textbf{97.8}} & \multicolumn{1}{c}{86.2}          & \multicolumn{1}{c|}{92.7}          & \multicolumn{1}{c}{90.1}\\
Jafarian et al.~\cite{jafarian2018monet}           & 98.6                     & \multicolumn{1}{c}{68.2}          & \multicolumn{1}{c}{38.3}          & \multicolumn{1}{c}{23.5}          & \multicolumn{1}{c}{28.9}          & \multicolumn{1}{c}{45.2}          & \multicolumn{1}{c|}{69.2}          & \multicolumn{1}{c||}{72.5}       
& 93.6                              & \multicolumn{1}{c}{64.5}          & \multicolumn{1}{c}{35.8}          & \multicolumn{1}{c}{24.5}          & \multicolumn{1}{c}{34.9} & \multicolumn{1}{c}{42.8}          & \multicolumn{1}{c|}{70.2}          & \multicolumn{1}{c}{70.8} \\\hline
\textbf{Ours}                      & \textbf{98.8}                     & \multicolumn{1}{c}{\textbf{93.1}} & \multicolumn{1}{c}{\textbf{78.5}} & \multicolumn{1}{c}{\textbf{66.8}} & \multicolumn{1}{c}{\textbf{98.5}} & \multicolumn{1}{c}{\textbf{98.3}} & \multicolumn{1}{c|}{\textbf{98.9}} & \multicolumn{1}{c||}{\textbf{95.6}} 
& \textbf{97.2}                     & \multicolumn{1}{c}{\textbf{88.3}} & \multicolumn{1}{c}{\textbf{68.3}} & \multicolumn{1}{c}{\textbf{52.4}} & \multicolumn{1}{c}{97.6} & \multicolumn{1}{c}{\textbf{89.3}} & \multicolumn{1}{c|}{\textbf{94.7}} & \multicolumn{1}{c}{\textbf{91.4}}
\\ \hline\hline
Human3.6M                   & Nec                      & \multicolumn{1}{c}{Sho}           & \multicolumn{1}{c}{Elb}           & \multicolumn{1}{c}{Wri}           & \multicolumn{1}{c}{Hip}           & \multicolumn{1}{c}{Kne}           & \multicolumn{1}{c|}{Ank}           & \multicolumn{1}{c||}{\textbf{AUC}} & Nec                      & \multicolumn{1}{c}{Sho}           & \multicolumn{1}{c}{Elb}           & \multicolumn{1}{c}{Wri}           & \multicolumn{1}{c}{Hip}           & \multicolumn{1}{c}{Kne}           & \multicolumn{1}{c|}{Ank}           & \multicolumn{1}{c}{\textbf{AUC}}                       
\\ \hline
Supervised  learning               & \multicolumn{1}{l}{92.1} & 75.3                              & 41.8                              & 26.5                              & 93.7                              & 82.5                              & 90.4                               & 86.2                                        

& \multicolumn{1}{l}{90.1}          & 76.3                              & 38.9                              & 20.8                              & \textbf{93.8}                     & 78.6                              & 83.2                               & 84.8         \\
Dong et al.~\cite{dong2018supervision}             & \multicolumn{1}{l}{95.4} & 88.6                              & 46.5                              & 35.2                              & 96.5                              & 95.6                              & 95.2                               & 91.6                              & \multicolumn{1}{l}{91.7}          & 81.4                              & 42.3                              & 25.6                              & 93.9                     & 83.4                              & 87.5                               & 86.9                          
\\
Jafarian et al.~\cite{jafarian2018monet}            & \multicolumn{1}{l}{95.8} & 50.8                              & 31.5                              & 18.5                              & 32.6                              & 40.8                              & 65.3                               & 69.9                        & \multicolumn{1}{l}{89.6}          & 48.3                              & 29.7                              & 20.5                              & 29.8                     & 34.9                              & 60.7                               & 65.2       \\\hline
\textbf{Ours }                     & \multicolumn{1}{l}{\textbf{97.9}} & \textbf{92.5}                     & \textbf{76.7}                     & \textbf{64.3}                     & 97.2                              & \textbf{97.6}                     & \textbf{96.9}                      & \textbf{94.8}                  & \multicolumn{1}{l}{\textbf{93.2}} & \textbf{92.8}                     & \textbf{67.3}                     & \textbf{49.6}                     & 93.7                     & \textbf{87.6}                     & \textbf{89.5}                      & \textbf{88.7}        \\ \hline
\end{tabular}
\end{center}
\vspace{-5mm}
\caption{We compare our approach with existing semi-supervised learning frameworks: (1) temporal supervision~\cite{dong2018supervision} and (2) cross-view supervision~\cite{jafarian2018monet}. We evaluate on two public human datasets (Panoptic Studio and Human3.6M) using PCKh measure. We test the generalizability by applying on unseen subjects. }
\label{tb:pckh_public}
\end{table*}

\begin{table*}[t]
\scriptsize
\begin{center}
\begin{tabular}{c|ccccccc|c||ccccccc|c}
\hline
\multicolumn{17}{c}{Monkey subject}\\
\hline
                    & \multicolumn{8}{c||}{DeepLabCut~\cite{Mathisetal2018}}                               & \multicolumn{8}{c}{Ours}                                                                                                     \\ \hline
\# annotations & Nose          & Hea  & Nec  & F.Leg & Paw  & Hip  & H. Leg & \textbf{AUC} & Nose          & Hea           & Nec           & F.Leg         & Paw           & Hip           & H. Leg        & \textbf{AUC}  \\ \hline
10                  & 92.1          & 93.5 & 90.6 & 59.4  & 28.2 & 97.3 & 63.2   & 73.9         & \textbf{93.2} & 94.6          & \textbf{91.4} & \textbf{83.2} & \textbf{43.9} & \textbf{92.1} & \textbf{85.5} & \textbf{89.1} \\
20                  & \textbf{95.9} & 95.7 & 95.2 & 68.3  & 30.8 & 98.3 & 70.1   & 78.7         & 95.1          & \textbf{99.3} & \textbf{98.7} & \textbf{95.2} & \textbf{50.1} & \textbf{93.5} & \textbf{95.7} & \textbf{92.2} \\
30                  & 95.3          & 95.8 & 96.7 & 73.7  & 33.2 & 98.5 & 75.6   & 80.3         & \textbf{95.4} & \textbf{99.1} & \textbf{98.5} & \textbf{95.9} & \textbf{54.8} & \textbf{95.7} & \textbf{96.0}   & \textbf{93.8} \\
40                  & \textbf{96.5} & 96.2 & 96.8 & 77.8  & 39.7 & 97.9 & 78.7   & 83.8         & 96.5          & \textbf{99.5} & \textbf{99.2} & \textbf{96.3} & \textbf{55.7} & \textbf{94.8} & \textbf{96.3} & \textbf{95.3} \\
50                  & 96.5          & 96.5 & 97.1 & 81.9  & 42.6 & 98.3 & 82.3   & 85.4         & \textbf{96.6} & \textbf{99.4} & \textbf{99.0}  & \textbf{96.4} & \textbf{56.3} & \textbf{95.1} & \textbf{96.7} & \textbf{96.2} \\ \hline
\hline
\multicolumn{17}{c}{Mouse subject}\\\hline
                  & \multicolumn{8}{c||}{DeepLabCut~\cite{Mathisetal2018}}   & \multicolumn{8}{c}{Ours}                                     \\ \hline
\# annotations & LF. paw & LH. paw & Tail & RF. paw & RH. paw & \multicolumn{1}{|c|}{\textbf{MAE}} & \textbf{RMSE} & \textbf{AUC} & LF. paw       & LH. paw       & Tail          & RF. paw       & RH. paw       & \multicolumn{1}{|c|}{\textbf{MAE}} & \textbf{RMSE} & \textbf{AUC}  \\ \hline
5              & 51.1    & 53.7    & 73.1 & 51.3    & 53.3    & \multicolumn{1}{|c|}{6.7}          & 8.7           & 63.5         & \textbf{57.6} & \textbf{58.5} & \textbf{76.6} & \textbf{57.9} & \textbf{58.1} & \multicolumn{1}{|c|}{\textbf{6.1}} & \textbf{8.4}  & \textbf{65.7} \\
10             & 60.0    & 61.9    & 78.5 & 60.6    & 61.1    & \multicolumn{1}{|c|}{5.8}          & 7.9           & 69.8         & \textbf{68.4} & \textbf{69.5} & \textbf{82.8} & \textbf{67.8} & \textbf{69.8} & \multicolumn{1}{|c|}{\textbf{4.9}} & \textbf{7.3}  & \textbf{73.6} \\
20             & 64.5    & 65.2    & 80.7 & 64.9    & 66.4    & \multicolumn{1}{|c|}{5.4}          & 7.7           & 74.2         & \textbf{73.9} & \textbf{75.6} & \textbf{85.4} & \textbf{74.5} & \textbf{75.0} & \multicolumn{1}{|c|}{\textbf{4.4}} & \textbf{6.5}  & \textbf{79.5} \\
40             & 67.3    & 67.1    & 82.1 & 66.7    & 67.3    & \multicolumn{1}{|c|}{5.0}          & 7.5           & 75.9         & \textbf{78.8} & \textbf{79.0} & \textbf{87.9} & \textbf{78.4} & \textbf{79.2} & \multicolumn{1}{|c|}{\textbf{3.9}} & \textbf{5.9}  & \textbf{81.4} \\ \hline
\end{tabular}
\end{center}
\vspace{-5mm}
\caption{We compare our approach with DeepLabCut~\cite{Mathisetal2018} that leverages a pre-trained model as varying the number of annotations. RMSE and MAE are measured in term of confidence map size ($46 \times 46$).}
\label{tb:pckh-deeplabcut}
\end{table*}

\begin{figure*}[t]
  \centering
  \subfigure[Human subject]{\includegraphics[height=0.2\textheight]{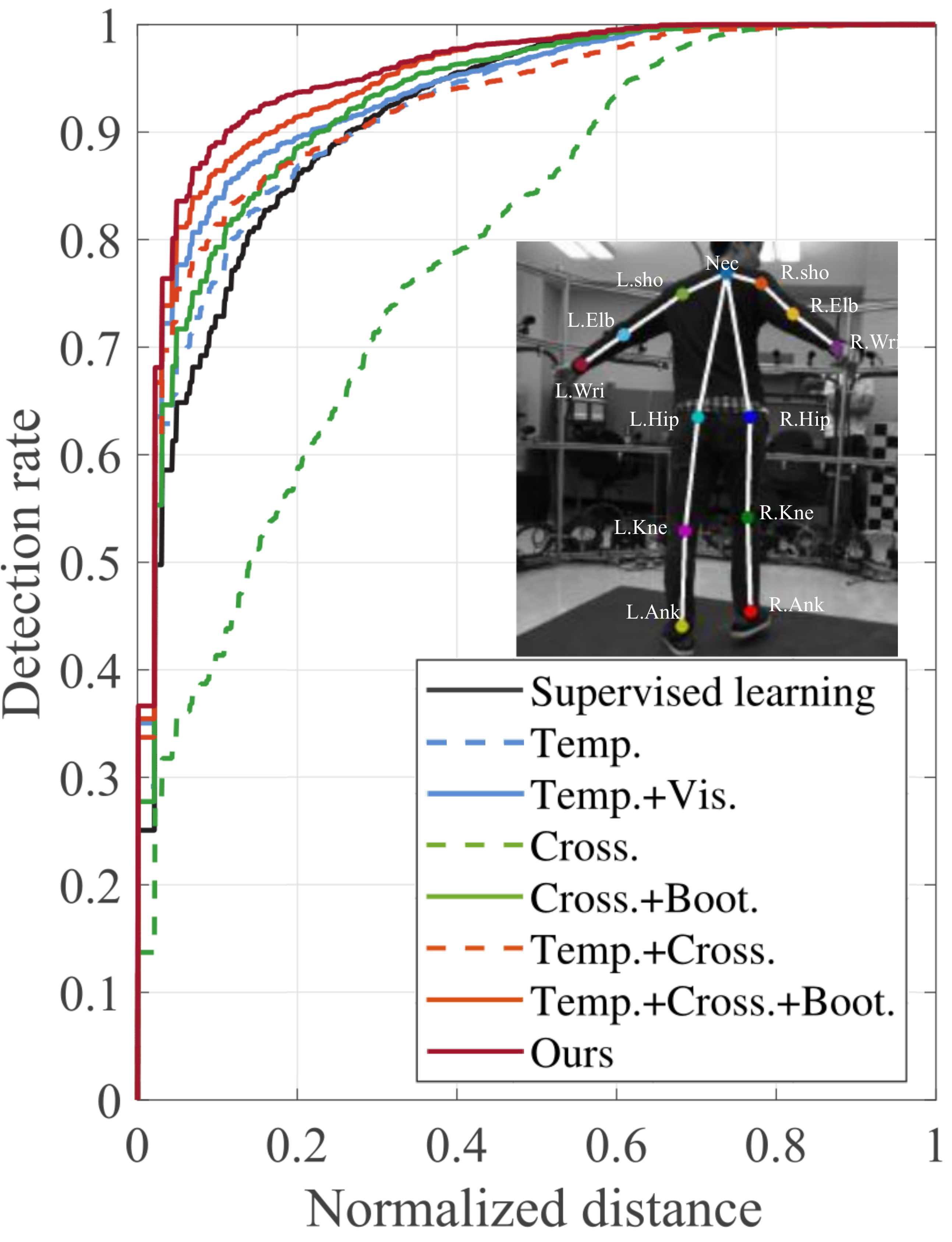}}~
  \subfigure[Dog subject]{\includegraphics[height=0.2\textheight]{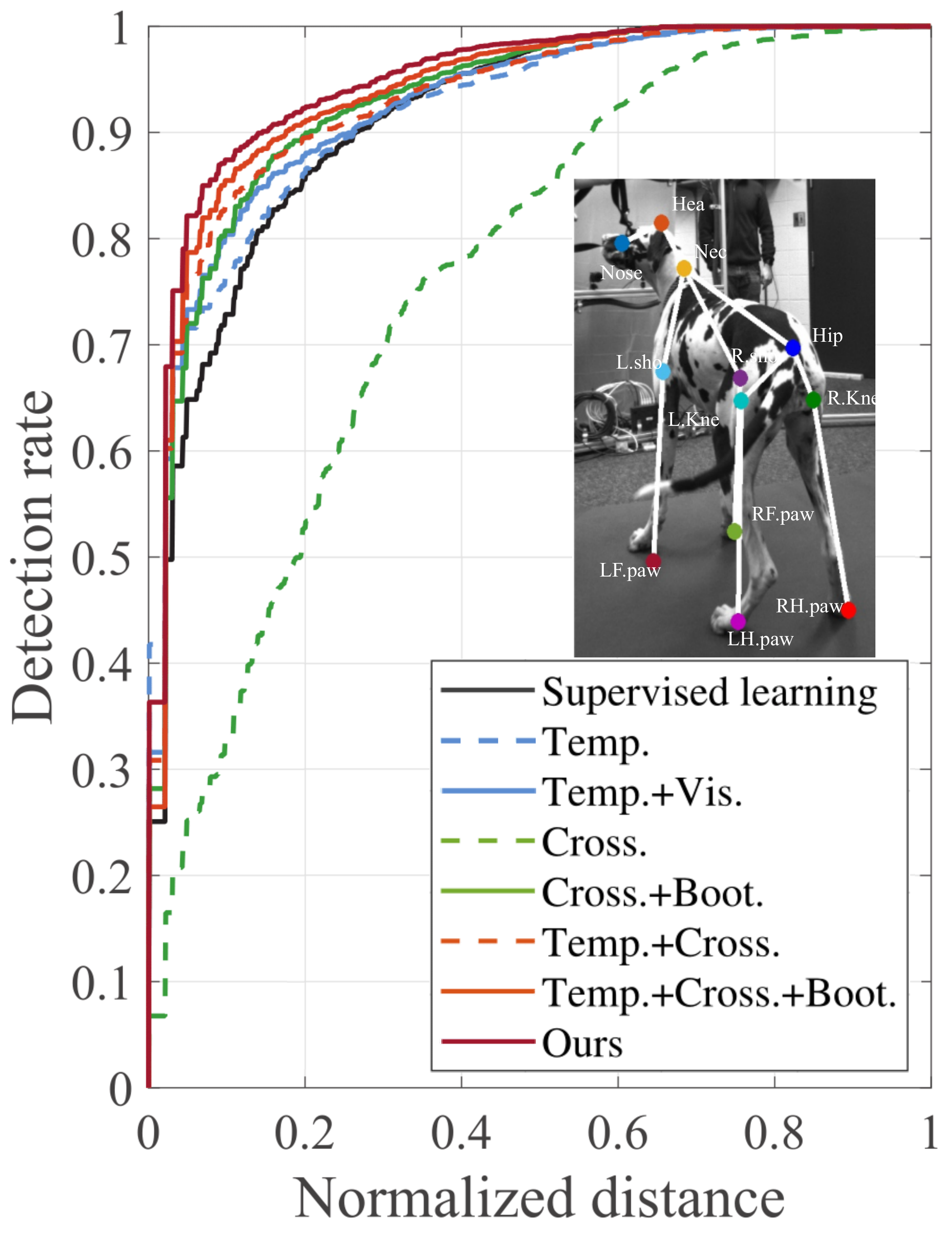}}~
  \subfigure[Monkey subject]{\includegraphics[height=0.2\textheight]{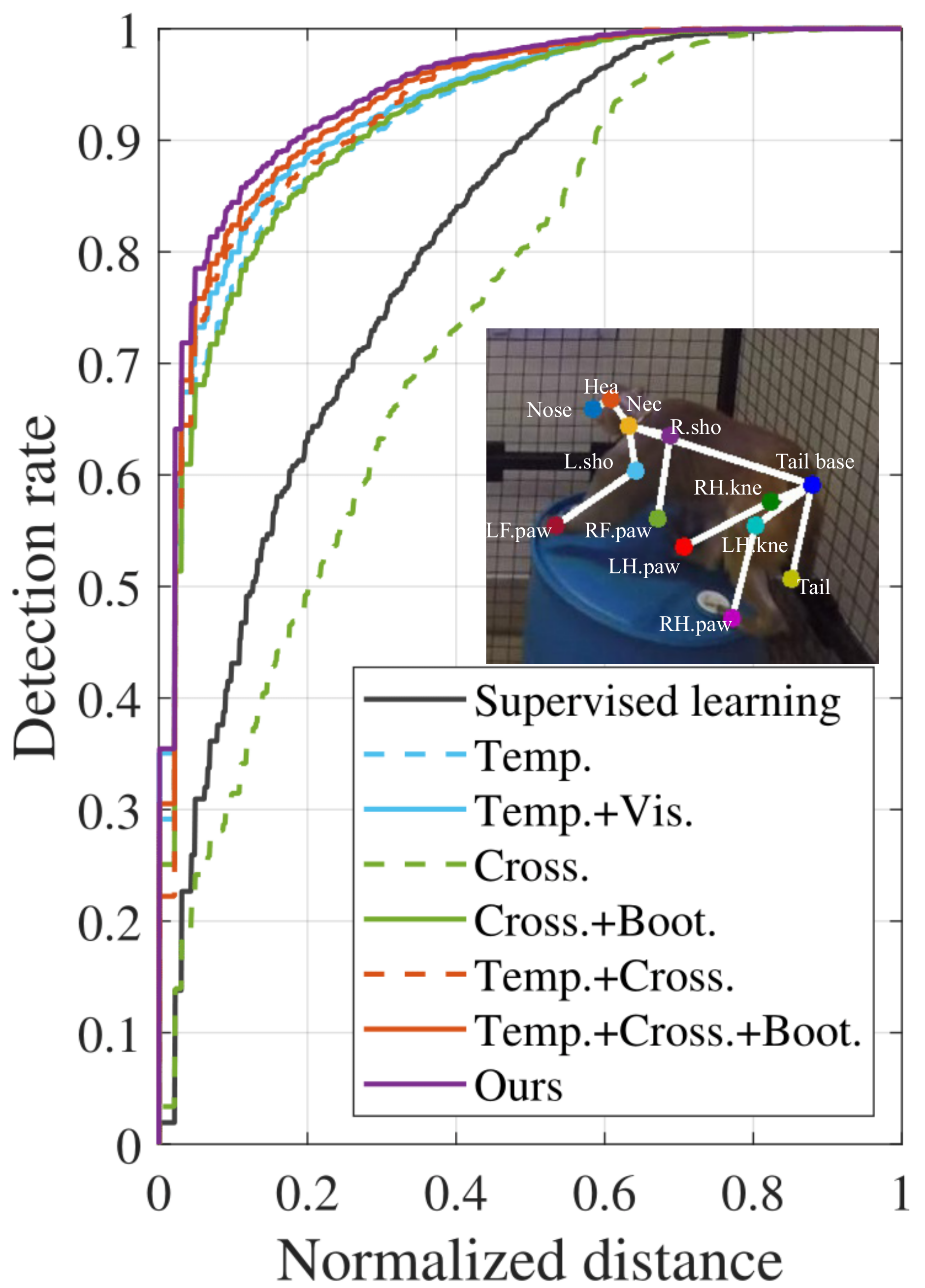}}~
  \subfigure[Monkey subject ]{\includegraphics[height=0.2\textheight]{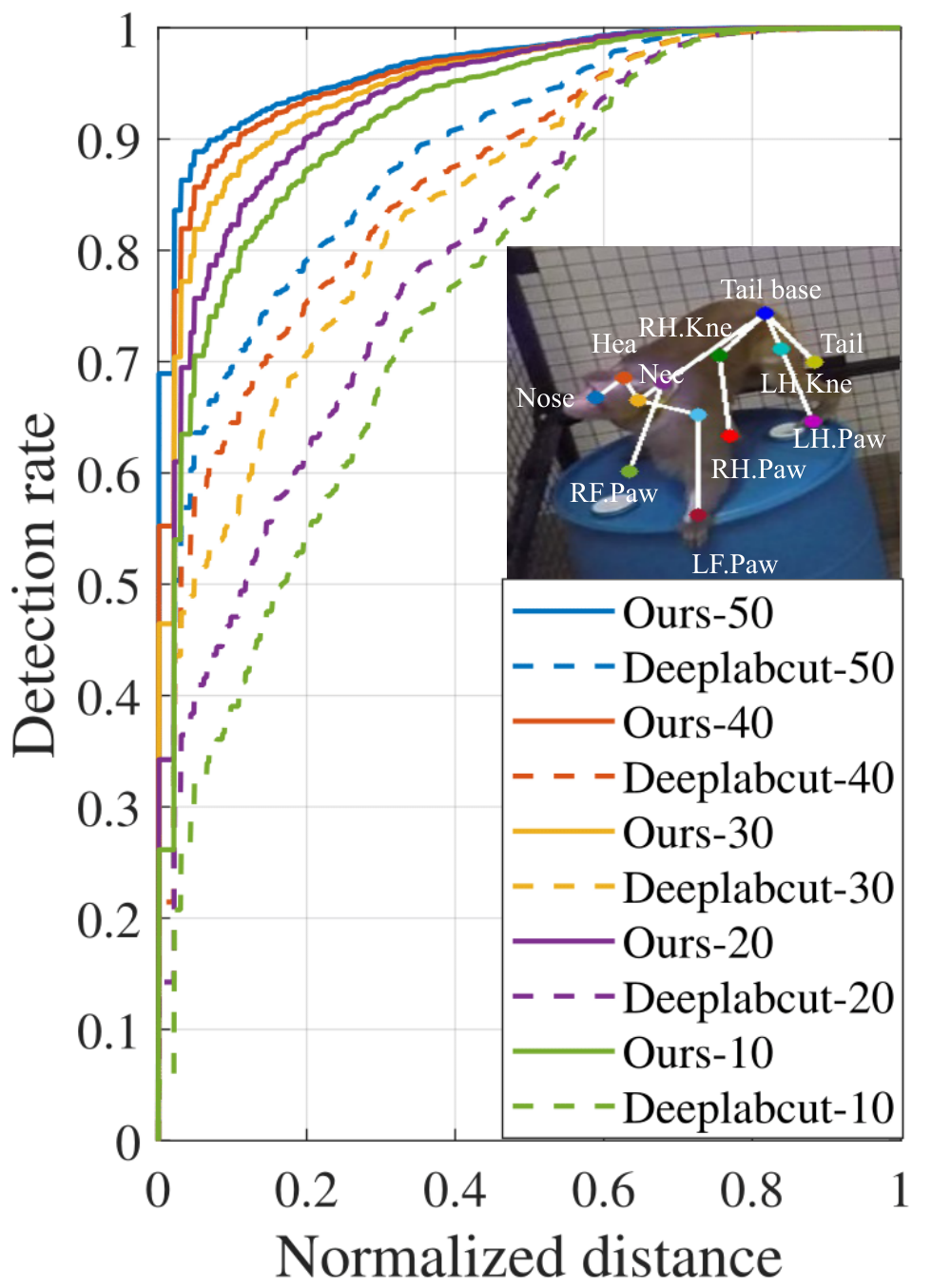}}~
  \subfigure[Mouse subject]{\includegraphics[height=0.2\textheight]{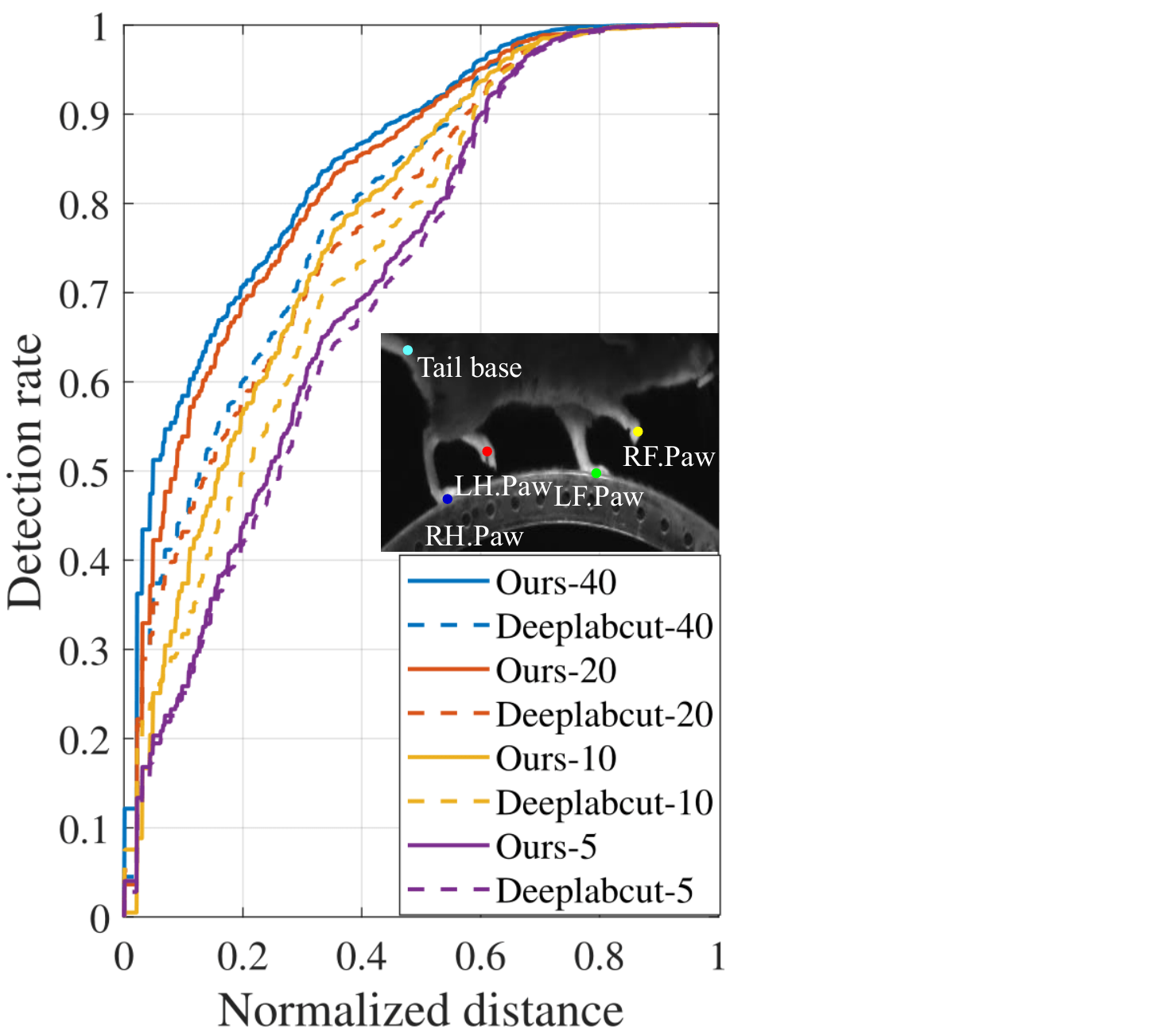}}
  \vspace{-3mm}
  \caption{(a-c) We conduct ablation study using a PCK measure on human, dog, and monkey subjects. (d-e) We compare Deeplabcut (Resnet 50)~\cite{Mathisetal2018} with ours on monkey and mouse subjects.}
  \label{fig:PCK_ablation}
  \vspace{-5mm}
\end{figure*}

\section{Experiments and Results}
\noindent\textbf{Datasets} We evaluate our approach using realworld multiview image streams of non-human and human species without a pre-trained model captured by multi-camera systems. (1) \textbf{Monkey subject} 35 cameras running at 60 fps are installed in a large cage ($9'\times12'\times9'$) that allows the free-ranging behaviors of monkeys. There are diverse monkey activities include grooming, hanging, and walking. The camera produces $1280\times 960$ images. The ground truth of keypoint and visibility is manually labeled. (2) \textbf{Dog subjects} Multi-camera system composed of 69 synchronized HD cameras (1024$\times$1280 at 30 fps) are used to capture the behaviors of multiple breeds of dogs including Dalmatian and Golden Retrievers. The ground truth is manually labeled. (3) \textbf{Mouse subject} We use a multiview mouse locomotion dataset used to evaluate DeepLabCut~\cite{Mathis:2018}. A single camera with a mirror generates multiview synchronized images of a head-fixed mouse running on a treadmill. The scene is captured at 200 Hz and the keypoints are fully annotated manually\footnote{The data were prepared by Rick Warren in Sawtell lab~\cite{Mathis:2018}.}. (4) \textbf{Human subject I} A multiview behavioral imaging system composed of 69 synchronized HD cameras capture human activities at 30 fps with 1024$\times$1280 resolution. We select 51 consecutive synchronized frames from 10 camera as training streams. Two end frames are used for the labeled data (20 images) and the rest images are used for the unlabeled data (490). The human pose detectors are used to triangulate the 3D pose to provide the ground truth. (5) \textbf{Human subject II} We test our approach on two publicly available datasets for human subjects: Panoptic Studio dataset~\cite{joo_iccv_2015} and Human3.6M~\cite{h36m_pami}. For the Panoptic Studio dataset, we use 31 HD videos ($1920\times1080$ at 30 Hz). The scenes includes diverse subjects with social interactions that introduce severe social occlusion. The Human3.6M dataset is captured by 4 HD cameras that includes variety of single actor activities, e.g., sitting, running, and eating/drinking.

\noindent\textbf{Metric} We use a measure of the probability of correct keypoint (PCK) and PCKh that accounts for 50\% of head length as a correct match. Area under curve (AUC) on PCK is also used to measure overall accuracy given fixed threshold. 

\noindent\textbf{Ablation Study} We conduct ablation study to analyze the effect of each component in our network. (1) supervised learning with the labeled data; (2) semi-supervised learning with temporal supervision; (3) temporal supervision + visibility supervision; (4) cross-view supervision; (5) cross-view supervision + visibility supervision + bootstrapping; (6) cross-view supervision + temporal supervision; (7) cross-view supervision + temporal supervision + bootstrapping; (8) ours (cross-view supervision + temporal supervision + visibility supervision + bootstrapping). Except for the fully supervised learning, all network designs utilizes the unlabeled data. 

Table~\ref{tb:pckh} and Figure~\ref{fig:PCK_ablation}(a-c) summarize the result of ablation study on human, dog, and monkey subjects. Our approach achieves $95.1\%$ on the Human dataset and $94.8\%$ AUC on the Dog dataset, which outperforms the other 2 unsupervised baselines, temporal supervision and cross-supervision, by $3.4\%$ and $16.4\%$ AUC respectively on the Human dataset, and by $2.8\%$ and $18.8\%$ AUC on the Dog dataset. In addition, visibility probability improves temporal supervision by $1.8\%$ AUC on the Human dataset and $2.65\%$ AUC on the Dog dataset. Similarly, data augmentation improves cross-view supervision by $16.6\%$ AUC on the Human dataset and $13.9\%$ AUC on the Dog dataset.

\noindent\textbf{Comparison with Semi-supervised Learning} We compare our approach with existing semi-supervised learning frameworks that use (1) temporal supervision~\cite{dong2018supervision} and (2) cross-view supervision~\cite{jafarian2018monet} on two publicly available human subject datasets (Panoptic Studio and Human3.6M). No pre-trained model is used for the comparison.

Table~\ref{tb:pckh_public} summarizes the PCKh measure of methods including fully supervised learning with the labeled data. Leveraging semi-supervised learning enhances the detection accuracy (there exists significant performance degradation of cross-view supervision due to long interval between the annotated frames). This shows that our approach leverages the unlabeled data better through the tight integration of temporal and cross-view supervisions. Also we test the generalizability of the trained pose detector by applying to the unseen subjects who are not used as unlabeled data. For Panoptic Studio, Dance 1 is used for the labeled and unlabeled and Dance 2 is used for the unseen data, and for Human3.6M, Eating and Discussion are used for the labeled and unlabeled data, and Greeting is used for the unseen data. The trend is similar to the unlabeled data, i.e., our approach shows stronger generalization power. 


\noindent\textbf{Comparison with DeepLabCut} We compare our approach with DeepLabCut~\cite{Mathisetal2018} that leverages a pre-trained model (ResNet 50~\cite{he:2016} trained on ImageNet~\cite{ILSVRC15}). In particular, we focus on non-human subjects (monkeys and mice) to reflect the strength of DeepLabCut. Two datasets differ in range of motion. For the mouse locomotion, the head of the mouse is stabilized where the range of motion is restricted to leg motion on the treadmill. On the other hand, the monkey activities are completely unconstrained, which produces severe self-occlusion and pose variation. 

Table~\ref{tb:pckh-deeplabcut} and Figure~\ref{fig:PCK_ablation}(d-e) summarize the performance comparison with respect to the number of annotations. A notable difference is that the performance gap of the monkey activities is much higher that that of the mouse, e.g., for 10 annotated data, our approach outperforms 15\% for the monkey and 3.5\% for the mouse. This indicates that our approach is more resilient to large appearance change induced by free-ranging activities.

\noindent\textbf{Qualitative Evaluation} We show the qualitative result in Figure~\ref{fig:teaser}. See Appendix and Supplementary Video for extensive qualitative result.

\section{Summary}
 We present a new semi-supervised learning framework to train a keypoint detector from multiview image streams. We integrate three self-supervisionary signals to effectively utilize a large amount of the unlabeled multiview data: (1) the cross-view supervision that enforces geometric consistency through the epipolar constraint across views; (2) the temporal supervision that constrains keypoint detection to be in accordance with dense optical flow; and (3) the visibility supervision that validates the detected keypoint in the presence of severe self-occlusion. We embed these supervisions into a new network design composed of three pathways in a differentiable fashion, allowing end-to-end training. We demonstrate that our approach outperforms existing semi-supervised learning approaches~\cite{dong2018supervision,jafarian2018monet} and DeepLabCut~\cite{Mathisetal2018} that uses a pre-trained model. The resulting network precisely detects the keypoints of both non-human and human subjects with highly limited labeled data ($<4\%$). 
 

 
 
 

{\small
\bibliographystyle{ieee}
\bibliography{egbib}

\begin{thebibliography}{10}\itemsep=-1pt

\bibitem{Andriluka:2014}
M.~Andriluka, L.~Pishchulin, P.~Gehler, and B.~Schiele.
\newblock 2d human pose estimation: New benchmark and state of the art
  analysis.
\newblock In {\em CVPR}, 2014.

\bibitem{arev:2014}
I.~Arev, H.~S. Park, Y.~Sheikh, J.~K. Hodgins, and A.~Shamir.
\newblock Automatic editing of footage from multiple social cameras.
\newblock {\em SIGGRAPH}, 2014.

\bibitem{baker2004lucas}
S.~Baker and I.~Matthews.
\newblock Lucas-kanade 20 years on: A unifying framework.
\newblock {\em IJCV}, 2004.

\bibitem{bertasius:2016_unsupervised}
G.~Bertasius, S.~X. Yu, H.~S. Park, and J.~Shi.
\newblock Exploiting visual-spatial first-person co-occurrence for
  action-object detection without labels.
\newblock In {\em ICCV}, 2017.

\bibitem{bolme2010visual}
D.~S. Bolme, J.~R. Beveridge, B.~A. Draper, and Y.~M. Lui.
\newblock Visual object tracking using adaptive correlation filters.
\newblock In {\em CVPR}, 2010.

\bibitem{Byravan:2016}
A.~Byravan and D.~Fox.
\newblock {SE3-nets}: Learning rigid body motion using deep neural networks.
\newblock In {\em ICRA}, 2016.

\bibitem{cao2016realtime}
Z.~Cao, T.~Simon, S.-E. Wei, and Y.~Sheikh.
\newblock Realtime multi-person 2d pose estimation using part affinity fields.
\newblock {\em CVPR}, 2016.

\bibitem{dong2018supervision}
X.~Dong, S.-I. Yu, X.~Weng, S.-E. Wei, Y.~Yang, and Y.~Sheikh.
\newblock Supervision-by-registration: An unsupervised approach to improve the
  precision of facial landmark detectors.
\newblock In {\em CVPR}, 2018.

\bibitem{Fischler:1981}
M.~A. Fischler and R.~C. Bolles.
\newblock Random sample consensus: A paradigm for model fitting with
  applications to image analysis and automated cartography.
\newblock {\em ACM Comm.}, 1981.

\bibitem{furukawa:2008}
Y.~Furukawa and J.~Ponce.
\newblock Dense 3d motion capture from synchronized video streams.
\newblock In {\em CVPR}, 2008.

\bibitem{hartley:2004}
R.~Hartley and A.~Zisserman.
\newblock {\em Multiple View Geometry in Computer Vision}.
\newblock Cambridge University Press, second edition, 2004.

\bibitem{he:2016}
K.~He, X.~Zhang, S.~Ren, and J.~Sun.
\newblock Deep residual learning for image recognition.
\newblock In {\em CVPR}, 2016.

\bibitem{henriques2015high}
J.~F. Henriques, R.~Caseiro, P.~Martins, and J.~Batista.
\newblock High-speed tracking with kernelized correlation filters.
\newblock {\em TPAMI}, 2015.

\bibitem{h36m_pami}
C.~Ionescu, D.~Papava, V.~Olaru, and C.~Sminchisescu.
\newblock Human3.6m: Large scale datasets and predictive methods for 3d human
  sensing in natural environments.
\newblock {\em TPAMI}, 2014.

\bibitem{jafarian2018monet}
Y.~Jafarian, Y.~Yao, and H.~S. Park.
\newblock Monet: Multiview semi-supervised keypoint via epipolar divergence.
\newblock {\em arXiv}, 2018.

\bibitem{joo_iccv_2015}
H.~Joo, H.~Liu, L.~Tan, L.~Gui, B.~Nabbe, I.~Matthews, T.~Kanade, S.~Nobuhara,
  and Y.~Sheikh.
\newblock Panoptic studio: A massively multiview system for social motion
  capture.
\newblock In {\em ICCV}, 2015.

\bibitem{joo_cvpr_2014}
H.~Joo, H.~S. Park, and Y.~Sheikh.
\newblock Map visibility estimation for large-scale dynamic 3d reconstruction.
\newblock In {\em CVPR}, 2014.

\bibitem{kanazawaHMR18}
A.~Kanazawa, M.~J. Black, D.~W. Jacobs, and J.~Malik.
\newblock End-to-end recovery of human shape and pose.
\newblock In {\em CVPR}, 2018.

\bibitem{Kullback:1951}
S.~Kullback and R.~A. Leibler.
\newblock On information and sufficiency.
\newblock {\em Annals of Mathematical Statistics}, 1951.

\bibitem{lin:2017}
C.-H. Lin and S.~Lucey.
\newblock Inverse compositional spatial transformer networks.
\newblock In {\em CVPR}, 2017.

\bibitem{lin:2014}
T.-Y. Lin, M.~Maire, S.~Belongie, J.~Hays, P.~Perona, D.~Ramanan,
  P.~Doll\`{a}r, and C.~L. Zitnick.
\newblock Microsoft coco: Common objects in context.
\newblock In {\em ECCV}, 2014.

\bibitem{liu2018two}
H.~Liu, J.~Lu, J.~Feng, and J.~Zhou.
\newblock Two-stream transformer networks for video-based face alignment.
\newblock {\em TPAMI}, 2018.

\bibitem{longuet-higgins:1981}
H.~C. Longuet-Higgins.
\newblock A computer algorithm for reconstructing a scene from two projections.
\newblock {\em Nature}, 1981.

\bibitem{Mathisetal2018}
A.~Mathis, P.~Mamidanna, K.~M. Cury, T.~Abe, V.~N. Murthy, M.~W. Mathis, and
  M.~Bethge.
\newblock Deeplabcut: markerless pose estimation of user-defined body parts
  with deep learning.
\newblock {\em Nature Neuroscience}, 2018.

\bibitem{Mathis:2018}
A.~Mathis and R.~A. Warren.
\newblock On the inference speed and video-compression robustness of
  deeplabcut.
\newblock In {\em bioRxiv}, 2018.

\bibitem{newell2016stacked}
A.~Newell, K.~Yang, and J.~Deng.
\newblock Stacked hourglass networks for human pose estimation.
\newblock In {\em ECCV}, 2016.

\bibitem{peng2016recurrent}
X.~Peng, R.~S. Feris, X.~Wang, and D.~N. Metaxas.
\newblock A recurrent encoder-decoder network for sequential face alignment.
\newblock In {\em ECCV}, 2016.

\bibitem{ILSVRC15}
O.~Russakovsky, J.~Deng, H.~Su, J.~Krause, S.~Satheesh, S.~Ma, Z.~Huang,
  A.~Karpathy, A.~Khosla, M.~Bernstein, A.~C. Berg, and L.~Fei-Fei.
\newblock {ImageNet Large Scale Visual Recognition Challenge}.
\newblock {\em IJCV}, 2015.

\bibitem{simon:2017}
T.~Simon, H.~Joo, I.~Matthews, and Y.~Sheikh.
\newblock Hand keypoint detection in single images using multiview
  bootstrapping.
\newblock In {\em CVPR}, 2017.

\bibitem{toshev2014deeppose}
A.~Toshev and C.~Szegedy.
\newblock Deeppose: Human pose estimation via deep neural networks.
\newblock In {\em CVPR}, 2014.

\bibitem{drcTulsiani17}
S.~Tulsiani, T.~Zhou, A.~A. Efros, and J.~Malik.
\newblock Multi-view supervision for single-view reconstruction via
  differentiable ray consistency.
\newblock In {\em CVPR}, 2017.

\bibitem{Velliste:2008}
M.~Velliste, S.~Perel, M.~C. Spalding, A.~S. Whitford, and A.~B. Schwartz.
\newblock Cortical control of a prosthetic arm for self-feeding.
\newblock {\em Nature}, 2008.

\bibitem{Vijayanarasimhan:2017}
S.~Vijayanarasimhan, S.~Ricco, C.~Schmid, R.~Sukthankar, and K.~Fragkiadaki.
\newblock Sfm-net: Learning of structure and motion from video.
\newblock In {\em arXiv}, 2017.

\bibitem{wei2016convolutional}
S.-E. Wei, V.~Ramakrishna, T.~Kanade, and Y.~Sheikh.
\newblock Convolutional pose machines.
\newblock In {\em CVPR}, 2016.

\bibitem{xiaolong:2015}
A.~G. Xiaolong~Wang.
\newblock Unsupervised learning of visual representations using videos.
\newblock In {\em ICCV}, 2015.

\bibitem{yoon20173d}
J.~S. Yoon, Z.~Li, and H.~S. Park.
\newblock 3d semantic trajectory reconstruction from 3d pixel continuum.
\newblock In {\em CVPR}, 2017.

\bibitem{zhou2017unsupervised}
T.~Zhou, M.~Brown, N.~Snavely, and D.~G. Lowe.
\newblock Unsupervised learning of depth and ego-motion from video.
\newblock In {\em CVPR}, 2017.

\end{thebibliography}
}

\end{document}